\documentclass[sigconf,nonacm]{acmart}
\setcopyright{none}
\renewcommand\footnotetextcopyrightpermission[1]{}

\usepackage{graphicx} 
\usepackage{amsmath}
\usepackage{amsfonts}
\usepackage{algorithm}
\usepackage{algpseudocode}

\newcommand{\authorsep}{\texorpdfstring{,\quad }{, }}

\title{SeqFold: Timestamp-Addressable Compressed States for User Action Sequence Backbones}
\author{Mianwei~Zhou\authorsep Duo~Zhang\authorsep Henry~Tsai\authorsep Yuqi~Zheng\authorsep Jian~Li\authorsep Juexin~Lin\authorsep Sujay~Khandagale\authorsep Zheng~Zhang\authorsep Vijay~Rajaram\authorsep Dong~Jiao\authorsep Tao~Cheng}
\email{{mizhou,duozhang,mitsai,yuqizheng,jili31,julin9,sukhanda,zhzhang2,vijay,dojiao,taocheng}@coupang.com}
\affiliation{%
  \institution{Coupang}
  \city{Mountain View}
  \country{United States}
}

\begin{document}

\begin{abstract}
Existing sequential user modeling methods mainly focus on improving the backbone encoder or learning task-specific user representations for a single prediction objective. We instead study how to export the multi-layer user states of a frozen long-history backbone as compact, timestamp-addressable representations for downstream training backfill and serving. In this setting, the user state at a given timestamp summarizes the user's actions before that time, so downstream systems need only the state aligned to each target timestamp rather than the full hidden-state timeline. To this end, we present \texttt{SeqFold}, a compressed-state interface that combines learned multi-channel fusion, channel-specific temporal decay, and Lookup-Free Quantization (LFQ) to produce compact discrete codes from frozen backbone states. Task-specific decoders then map these shared codes into flexible embedding spaces for downstream models. To make timestamp-addressable export practical at scale, we further develop a fused streaming implementation that updates the fusion recurrence after every action while materializing states only at target timestamps. On large-scale proprietary datasets, SeqFold reduces per-state representation storage by $3{,}400\times$ while preserving most of the downstream ranking performance of the uncompressed model, showing that SeqFold is an effective interface for reusing long-history sequence backbones in downstream models.
\end{abstract}



\maketitle
\section{Introduction}
Large-scale e-commerce platforms increasingly rely on user behavior sequence models to capture long-term preferences, evolving purchase intent, and cross-surface behavior patterns. A user action sequence can contain thousands of heterogeneous actions, such as search queries, product impressions, clicks, add-to-cart events, and purchases. Transformer-based architectures are well suited to this setting because they can jointly encode semantic, contextual, and temporal dependencies across long user histories~\cite{kang2018sasrec, sun2019bert4rec, li2020tisasrec}. In practical large-scale systems, these sequence models often serve as reusable user representation backbones.

\begin{figure}[t]
\centering
\makebox[\linewidth][c]{\hspace*{-0.02\linewidth}\includegraphics[width=1.1\linewidth]{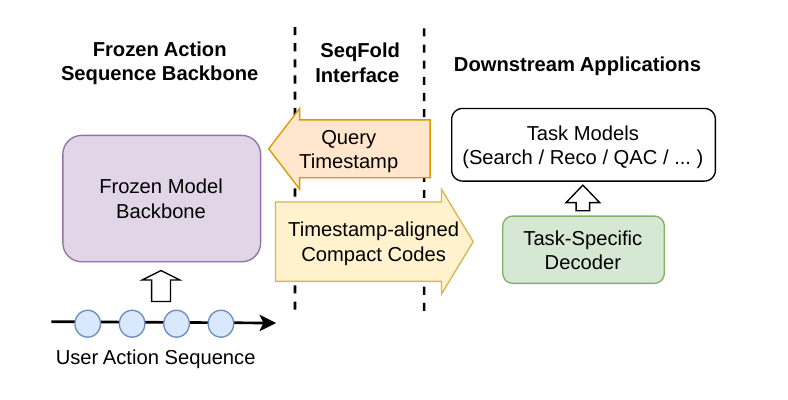}}
\Description{Overview of SeqFold as an interface between a frozen action sequence backbone and downstream applications. User action sequences are encoded by a frozen model backbone, downstream applications query SeqFold with timestamps, SeqFold returns timestamp-aligned compact codes, and task-specific decoders map those codes for task models such as search, recommendation, and query autocomplete.}
\vspace{-4mm}
\caption{SeqFold as a timestamp-addressable interface between a frozen action sequence backbone and downstream applications. Given a target timestamp, SeqFold returns timestamp-aligned compact codes that downstream tasks decode with task-specific decoders for downstream training backfill and serving.}
\vspace{-5mm}
\label{fig:intro_framework}
\end{figure}

While this backbone reuse is attractive, it creates a mismatch between what the upstream sequence model produces and what downstream systems actually need. A frozen long-history backbone can produce rich multi-layer states over a user's full action sequence, but downstream applications do not directly need that entire hidden-state timeline. Instead, for downstream ranking, both offline backfill and online serving need only the user state aligned to each target timestamp, that is, a summary of the user's actions before that time. Passing the raw backbone states through this interface is nevertheless impractical. In our main experiments, each user sequence contains at most 2,000 actions and the backbone uses 4 Transformer layers with hidden dimension 256, yielding up to approximately $2.05 \times 10^6$ floating-point values in the full hidden-state timeline for a single user sequence. Delivering such a high-dimensional payload to downstream models would create severe storage, transmission, and memory costs.

To address this bottleneck, we study the problem of \textit{timestamp-addressable compressed-state export for downstream training backfill and serving}. Given a user history and a target timestamp, the goal is to export a compact representation of the user state available before that time. As illustrated in Figure~\ref{fig:intro_framework}, our focus is not on accelerating the backbone itself, but on the interface between an upstream frozen long-history backbone and downstream consumers that need compact timestamp-aligned user states. We therefore propose \texttt{SeqFold}, a timestamp-addressable compressed-state interface for this setting. Given a target timestamp, \texttt{SeqFold} returns compact codes derived from the frozen backbone and aligned to that timestamp. Task-specific decoders then map these shared codes into continuous representations for downstream models. SeqFold instantiates this interface with learned multi-channel fusion, channel-specific temporal decay, and Lookup-Free Quantization (LFQ)~\cite{yu2024language}, producing compact discrete codes for flexible downstream use.

A timestamp-addressable interface changes not only what must be stored, but also what must be materialized during training backfill and serving. Let $B$ denote batch size, $N$ sequence length, $D$ hidden dimension, and let $C$ denote the number of fusion channels in \texttt{SeqFold}. A direct tensorized implementation of \texttt{SeqFold}'s action-level recurrence would construct a channel-specific state for every action prefix, expanding the activation footprint from $[B,N,D]$ to $[B,N,C,D]$ and leading to severe memory pressure. Yet the target workload does not require this dense prefix timeline: it needs only the latest causally accessible state at each target timestamp. We therefore develop a fused streaming implementation that scans each sequence chronologically, updates the fusion recurrence after every action, and materializes states only at target timestamps. During compressor training, periodic export points allow nearby target timestamps to share the latest causally valid exported state while improving efficiency.

In summary, the primary contributions of this paper are as follows:
\begin{itemize}
\item We formulate downstream reuse of long-history backbone states as \textbf{timestamp-addressable compressed-state export}, where downstream training backfill and online serving consume a compact causal state aligned to each target timestamp rather than a single static user embedding.
\item We propose \texttt{SeqFold}, which instantiates this interface with \textbf{learned multi-channel fusion}, temporal decay, and LFQ encoding to compress multi-layer user representations into compact discrete codes.
\item We develop a \textbf{fused streaming export mechanism} that decouples state update from state materialization, preserving action-level recurrence while avoiding dense prefix-state tensors.
\item We show on proprietary datasets that SeqFold provides a strong \textbf{quality--storage trade-off} and that its timestamp-state generation scales efficiently with sequence length and export sparsity.
\end{itemize}

\section{Related Work}

\subsection{Sequential User Modeling and Long-History Retrieval}
Sequential recommendation models encode user histories to predict future actions. Early systems relied on recurrent or convolutional architectures such as GRU4Rec~\cite{hidasi2016gru4rec} and Caser~\cite{tang2018caser}. Later work introduced target-aware interest modeling, including DIN~\cite{zhou2018din} and DIEN~\cite{zhou2019dien}, while Transformer-based backbones such as SASRec~\cite{kang2018sasrec}, BERT4Rec~\cite{sun2019bert4rec}, and TiSASRec~\cite{li2020tisasrec} improved long-range dependency modeling. For very long histories, industrial systems often combine sequence modeling with retrieval or multi-interest summarization, for example SIM~\cite{pi2020sim} and MIND~\cite{ma2019mind}. These methods mainly improve the backbone encoder or produce task-specific user representations for a particular prediction objective. In contrast, SeqFold assumes a frozen backbone and studies how to export its multi-layer causal states as timestamp-addressable compressed states for downstream reuse.

\subsection{Reusable User Representations for Downstream Tasks}
A related line of work studies how upstream representation models can serve multiple downstream retrieval or ranking stages. Deep structured semantic models~\cite{huang2013dssm} and large-scale recommender stacks such as YouTube's two-stage architecture~\cite{covington2016deep} motivate the practical value of shared user representations, but they typically expose a single static or request-time embedding rather than a causal state aligned to an arbitrary target timestamp. SeqFold is closest in spirit to this reuse setting, but it exports compact states indexed by timestamp so that offline backfill and online serving can both consume a causal state aligned to the target time without replaying or storing the full hidden-state timeline.

\subsection{Long-Context Efficiency and State Summarization}
A separate line of work reduces the cost of processing long contexts within the backbone itself. Efficient attention methods such as Linformer~\cite{wang2020linformer}, Longformer~\cite{beltagy2020longformer}, and FlashAttention~\cite{dao2022flashattention} improve the asymptotic or practical cost of attention, while recurrence- and memory-based variants such as Transformer-XL~\cite{dai2019transformerxl} and Compressive Transformer~\cite{rae2020compressive} carry information across longer horizons. Other approaches summarize or compress long sequences inside the model, for example Perceiver IO~\cite{jaegle2021perceiverio}, TokenLearner~\cite{ryoo2021tokenlearner}, VQL~\cite{li2025vql}, and SlimPer~\cite{wang2026slimper}. These approaches are relevant because SeqFold also targets long histories, but they optimize a different stage of the pipeline: they redesign the backbone or its intermediate computation, whereas SeqFold operates after a frozen backbone has produced hidden states and focuses on storing and serving timestamp-aligned compressed states.

\subsection{Discrete Representation Learning for Compact State Interfaces}
Discrete latent methods provide tools for aggressively compressing continuous representations. VQ-VAE~\cite{oord2017vqvae} and VQ-VAE-2~\cite{razavi2019vqvae2} learn discrete codebooks for neural latent spaces, product quantization~\cite{jegou2011product} compresses vectors for large-scale similarity search, and Lookup-Free Quantization (LFQ)~\cite{yu2024language} replaces explicit codebook lookup with direct sign-based indexing. SeqFold applies LFQ in a different regime: it serializes multi-layer user states into compact causal codes that remain addressable by timestamp, while allowing each downstream system to decode the same shared codes into task-specific continuous spaces.

\section{Problem Setup and Interface Requirements}
\label{sec:background}
\subsection{Frozen User Action Sequence Backbone}

\begin{figure}[htbp] 
\centering 
\makebox[\linewidth]{\includegraphics[width=1.2\linewidth]{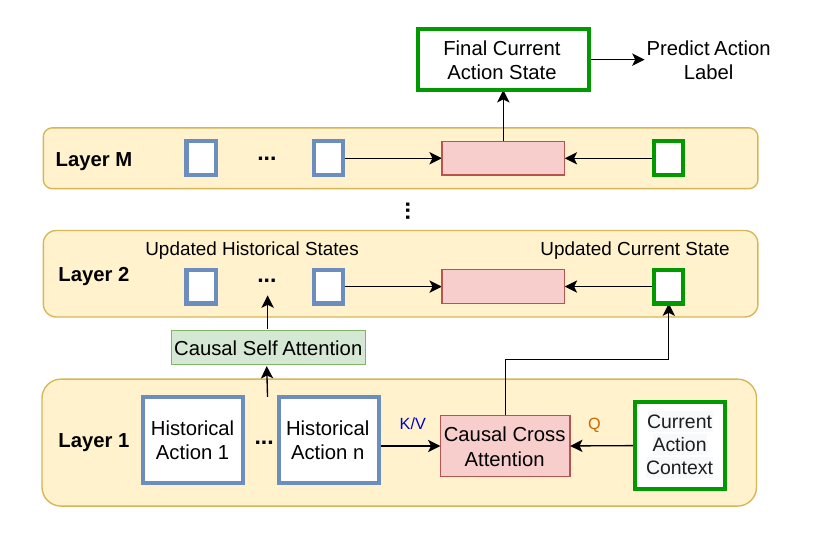}}
\Description{A two-stream user action sequence backbone. Historical action states are updated layer by layer by causal self-attention, while the current action context attends only to preceding historical states through causal cross-attention. The final current-action state is used to predict the action label.}
\caption{Two-stream action sequence backbone. At each layer, historical action states are updated by causal self-attention, while the current action context attends only to preceding historical states through causal cross-attention. The resulting current-action state is propagated upward through the stacked layers, and the final current-action state is used to predict the action label.}
\label{fig:action_sequence_model} 
\end{figure}

We assume a frozen causal user action sequence backbone that is trained on historical action data and refreshed periodically. Each action combines heterogeneous contextual and behavioral signals, and the backbone maps every action to a hidden state while respecting chronological causality. As illustrated in Figure~\ref{fig:action_sequence_model}, historical actions are updated layer by layer through causal self-attention, while the current action context queries only the preceding history through causal cross-attention to produce the final current-action state for label prediction. In our main experiments, the backbone stacks 4 Transformer layers with a 256-dimensional hidden space, and each user sequence contains at most 2,000 actions.

To train this backbone, we adopt a two-stream setup inspired by XLNet~\cite{yang2019xlnet}. Let $a_i$ denote the $i$-th action in a user's chronological sequence. We decouple each action into a \textit{context part} and a \textit{label part}, and predict the current action label from its concurrent context together with the historical actions that precede it:
\[P\!\left(\operatorname{Label}(a_{n+1}) \mid
\operatorname{Context}(a_{n+1}),
\operatorname{Label}(a_{1:n}),
\operatorname{Context}(a_{1:n})\right).\]
In the rest of the paper, we use $a_i$ for the action itself and $A_i^{\mathrm{full}}$ and $A_i^{\mathrm{ctx}}$ for the corresponding backbone-produced hidden states under the full-information and context-only views, respectively.
In practice, the main target is action-type prediction, and additional auxiliary objectives share the same backbone. The goal of this paper is not to redesign that backbone, but to expose its multi-layer causal states to downstream systems through a compact timestamp-addressable interface.

\subsection{Downstream Records and Timestamp-Addressable Export}
Downstream applications such as search, recommendation, and query autocomplete do not need the backbone's full hidden-state timeline. Instead, each offline training example or online serving request needs a compact causal state aligned to its target timestamp. A historical record must use only the user state available before its target event, and an online request must consume the latest causally valid state at request time.

Exporting the raw stacked hidden states is nevertheless impractical at this interface. For a history length $N$, $M$ backbone layers, and hidden dimension $D$, one user's complete multi-layer state contains $N \times M \times D$ floating-point values. In our main experiments, $N \le 2{,}000$, $M=4$, and $D=256$, so one user's complete multi-layer state can contain up to $2.048 \times 10^6$ floating-point values. For many lightweight downstream models, a standard record contains only a few hundred dense and sparse features, so injecting the full backbone state would overwhelm storage, transmission, and training-time memory budgets. The desired interface must therefore satisfy three requirements: it must be causal with respect to target time, compact enough for large-scale storage and delivery, and reusable across both backfill and online serving.

\section{SeqFold as a Compressed State Interface}
\label{sec:architecture}
\subsection{Architecture Overview}
\begin{figure*}[htbp]
\centering
\includegraphics[width=0.9\textwidth]{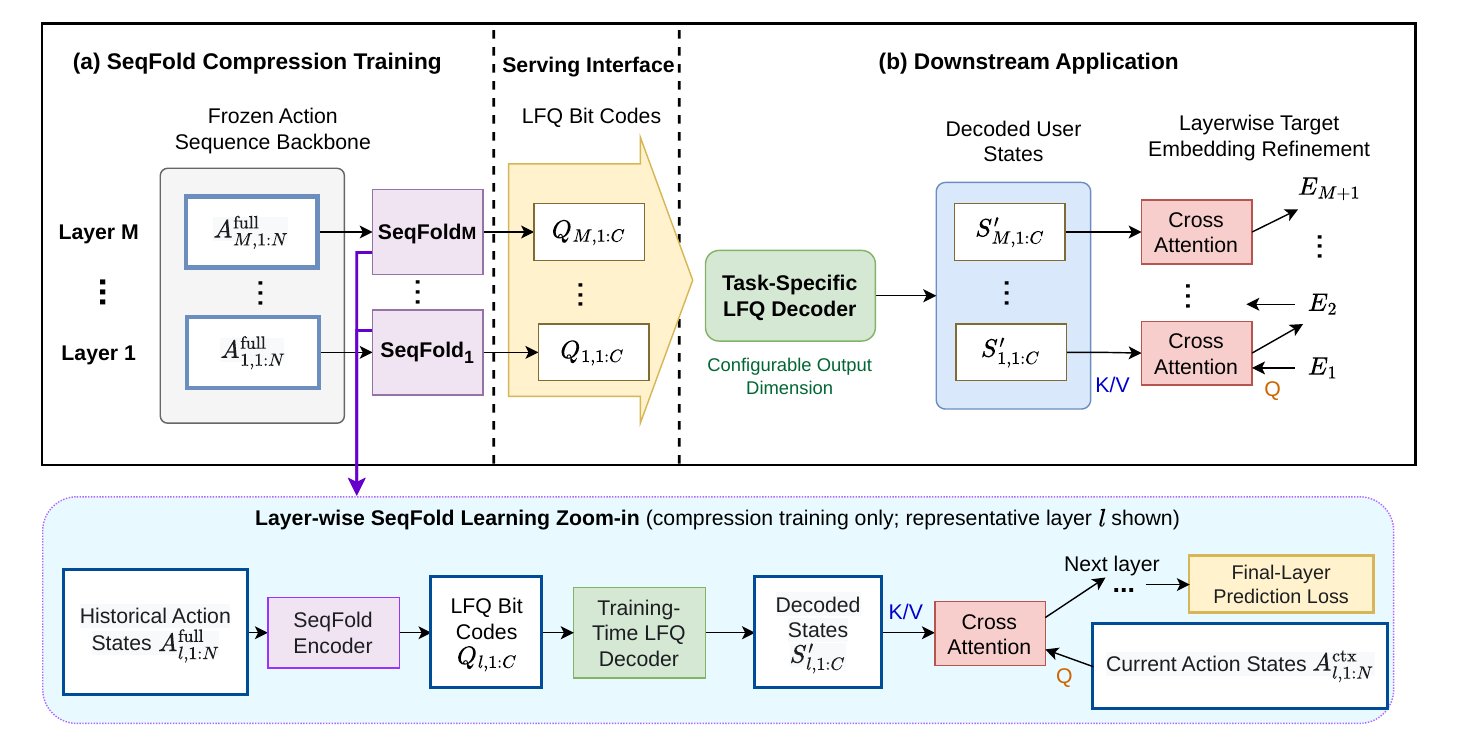}
\Description{An overview of SeqFold training and serving. A frozen multi-layer action sequence backbone produces layer-wise history states, SeqFold compresses them into shared LFQ codes, training uses a decoder and cross-attention path for supervision, and downstream applications reuse the shared codes with task-specific decoders during serving.}
\caption{SeqFold compression training and downstream serving. A frozen
multi-layer action sequence backbone first produces layer-wise history states,
which SeqFold compresses into shared LFQ bit codes. During compression
training, a training-only layer-wise learning path (bottom zoom-in; one
representative layer shown) decodes the compressed states and applies
cross-attention to the current action states to supervise the compressor with
the upstream prediction loss. During serving, each downstream application
reuses the shared codes and learns its own task-specific LFQ decoder with a
configurable output dimension for layer-wise cross-attention.}
\label{fig:seqfold_pipeline}
\end{figure*}


\texttt{SeqFold} sits between a frozen long-history action sequence backbone and the downstream systems that consume its states. The backbone first computes layer-wise history states, and SeqFold compresses those states after the backbone instead of rewriting the backbone attention blocks. It then exports a compact causal state indexed by target timestamp in place of the full per-action hidden-state timeline. For any requested time $t$, it summarizes only the actions before $t$ into reusable codes, enabling backfill and online serving to consume the same timestamp-addressable interface without replaying or storing the full continuous sequence. As illustrated in Figure~\ref{fig:seqfold_pipeline}, this interface has two operational phases: offline compression training and downstream serving. The bottom zoom-in further expands the training-only layer-wise learning path for one representative backbone layer.

During the \textbf{Training Phase}, the parameters of the pre-trained action sequence backbone are frozen, and \texttt{SeqFold} is trained to map the backbone history states into compact discrete codes through Lookup-Free Quantization (LFQ)~\cite{yu2024language}. The compressed states are then passed through a training-time LFQ decoder and a layer-wise cross-attention module. This training-only path is supervised with the same upstream action-prediction objective used to train the backbone and is used only to supervise the compressor; it is not the task-specific decoder used during downstream serving. The output of this phase is a compact matrix of quantized indices that summarizes the user behavior profile across multiple backbone layers.

During the \textbf{Downstream Serving Phase}, downstream models consume these shared codes at the requested serving or training timestamp rather than the original continuous hidden states. Each downstream task then learns its own task-specific LFQ decoder that maps the LFQ codes into a continuous space layer-by-layer before cross-attention scoring against its task-specific targets. The decoded embedding dimension need not match the original backbone dimension, allowing the same exported codes to support different downstream capacity budgets.

To illustrate the compression efficiency, consider a sequence with average history length $\bar N$, $M$ backbone layers, hidden dimension $D$, $C$ fusion vectors per layer, and $b_{\mathrm{LFQ}}$ bits per code. The raw, uncompressed user representation at one target timestamp requires $\bar N \times M \times D$ floating-point values, whereas \texttt{SeqFold} requires only $M \times C \times b_{\mathrm{LFQ}} / 8$ bytes after quantization. In our main experiments, $\bar N\approx 1{,}700$, $M=4$, $D=256$, $C=64$, and $b_{\mathrm{LFQ}}=64$, so the raw representation requires $1{,}700 \times 4 \times 256 = 1{,}740{,}800$ floating-point values, while \texttt{SeqFold} requires only $4 \times 64 \times 64 / 8 = 2{,}048$ bytes. The resulting gap can be substantial even for moderate layer counts and hidden dimensions, making compact timestamp-addressable export attractive for downstream consumption.

In the subsequent sections, we first describe one concrete instantiation of this timestamp-addressable interface. We then analyze the bottlenecks that arise from dense tensor materialization and present a fused streaming implementation that supports sparse timestamp export in long-history settings.

\subsection{A Concrete Instantiation}
\label{sec:initial_seqfold}
We next describe one concrete instantiation of the \texttt{SeqFold} interface. Its core mechanism adaptively gates, decays, and accumulates variable-length sequence embeddings into a fixed number of compressed discrete latent vectors. Table~\ref{tab:seqfold_notation} summarizes the key notation used below.

\begin{table}[t]
\small
\centering
\caption{Key notation used in the SeqFold instantiation.}
\label{tab:seqfold_notation}
\begin{tabular}{@{}lp{0.6\columnwidth}@{}}
\hline
\textbf{Symbol} & \textbf{Meaning} \\
\hline
$A_i^{\mathrm{full}}$ & Full-information hidden state of action $i$ \\
$A_i^{\mathrm{ctx}}$ & Context-only representation for action $i$ \\
$c$ & Fusion channel index \\
$\lambda_c$ & Temporal decay rate for channel $c$ \\
$b_{\mathrm{LFQ}}$ & Number of bits in each LFQ code \\
$S_{i,c}$ & Continuous fused state after the first $i$ actions in channel $c$ \\
$Q_{i,c}$ & LFQ code after the first $i$ actions in channel $c$ \\
\hline
\end{tabular}
\end{table}

Formally, for a sequence of length $N$, SeqFold computes gated action contributions $V_{i,c}$, continuous fusion states $S_{i,c}$, binary LFQ codes $Q_{i,c} \in \{0,1\}^{b_{\mathrm{LFQ}}}$, and a cross-attended representation $\hat{A}_{i+1}^{\mathrm{ctx}}$ as follows:
\begin{align}
V_{i,c} &= \sigma\left(f_c\left(A_i^{\mathrm{full}}\right)\right) \cdot A_i^{\mathrm{full}}, \label{eq:seqfold_v} \\
S_{i,c} &= \sum_{k\leq i} \left(e^{-\lambda_c(t_i-t_k)} V_{k,c}\right), \label{eq:seqfold_s} \\
Q_{i,c} &= \operatorname{LFQEncode}\left(\operatorname{RMSNorm}\left(S_{i,c}\right)\right), \label{eq:seqfold_q} \\
\hat{A}_{i+1}^{\mathrm{ctx}} &= \operatorname{Attn}\left(A_{i+1}^{\mathrm{ctx}}, \operatorname{LFQDecode}\left(Q_{i,1:C}\right)\right). \label{eq:seqfold_attn}
\end{align}
Here $\hat{A}_{i+1}^{\mathrm{ctx}}$ denotes the cross-attended downstream representation obtained after the context-only query representation attends to the decoded compressed history.

\subsubsection{Action Gating and Time-Aware Fusion}
As expressed in Equation~\ref{eq:seqfold_v}, the compression pipeline begins with dynamic gating. The scalar gate $\sigma\left(f_c\left(A_i^{\mathrm{full}}\right)\right)$ acts as a data-dependent routing weight that determines the contribution of the $i$-th action to the $c$-th fusion space. Because these channel gates are parameterized independently with sigmoids rather than normalized with a softmax across channels, a single action can contribute to multiple fusion channels when it contains information relevant to multiple semantic aspects.

To capture temporal dynamics in long-term user behaviors, these gated representations $V_{i,c}$ are aggregated across the entire historical horizon to construct the raw fusion vectors $S_{i,c}$, as detailed in Equation~\ref{eq:seqfold_s}. We introduce a time-dependent exponential decay factor $e^{-\lambda_c(t_i-t_k)}$, where $t_k$ and $t_i$ denote the respective action timestamps, and $\lambda_c$ is a channel-specific, learnable decay coefficient. This decay mechanism lets more recent actions contribute more strongly to the current state, while historical actions are discounted according to elapsed time. By condensing a sequence of thousands of historical actions into a fixed set of $C$ fusion vectors, SeqFold reduces downstream cross-attention complexity from $O(N)$ to $O(C)$, where $C \ll N$.

\subsubsection{Fusion Vector Compression and Downstream Decoding}
To reduce storage and network transmission cost, the continuous fusion vectors $S_{i,c}$ undergo standardization and quantization as formalized in Equation~\ref{eq:seqfold_q}. We first apply Root Mean Square Normalization ($\operatorname{RMSNorm}$) to stabilize the variance across the learned representation channels. The normalized vectors are then passed to a Lookup-Free Quantization ($\operatorname{LFQ}$) module, which maps continuous vectors directly to discrete binary codes via a sign-based bit assignment strategy. The resulting discrete representation $Q_{i,c} \in \{0,1\}^{b_{\mathrm{LFQ}}}$ can be serialized compactly; in the reference setting used in our experiments, $b_{\mathrm{LFQ}}=64$, so each code is stored as one 64-bit unsigned integer.

During downstream consumption (Equation~\ref{eq:seqfold_attn}), the target application loads the compact LFQ codes $Q_{i,1:C}$ that summarize the state after the first $i$ actions, so the next target representation $A_{i+1}^{\mathrm{ctx}}$ attends only to causally preceding history. These codes are mapped back to continuous space using an $\operatorname{LFQDecode}$ network, whose output dimension can be chosen to match the downstream model. The downstream model then uses a cross-attention layer ($\operatorname{Attn}$) so that its task-specific targets can query the reconstructed user behavior profile.

\section{Efficient Timestamp-State Generation}

The compressed-state interface above determines which states downstream records address, but it also changes how those states should be materialized. This section focuses on the systems realization of timestamp-addressable export.

\subsection{Bottlenecks of Dense State Materialization}
\label{sec:dense_materialization}
The formulation of \texttt{SeqFold} in Section~\ref{sec:initial_seqfold} defines a causal compressed state after every action prefix, so every historical action must update the running state. A direct tensorized implementation of this recurrence, however, can be prohibitively expensive because it materializes all prefix states explicitly. Let $B$ denote the training batch size, $N$ the historical sequence length, $D$ the action embedding dimension, and $C$ the number of fusion channels. The hidden states from the action backbone have shape $[B,N,D]$. A naive implementation of Equations~\ref{eq:seqfold_v}--\ref{eq:seqfold_s} expands each action representation across the fusion channels and materializes all prefix states in a tensor with shape $[B,N,C,D]$. In the long-history scalability benchmark studied later ($N=10^4$, $D=256$, and $C=64$), this tensor alone contains $10^4 \times 64 \times 256$ floating-point values per sample per layer, requiring approximately 0.66 GB in FP32 before accounting for other activations or backward storage.

Yet backfill and online serving do not need this dense prefix timeline. Downstream records request states only at target timestamps, each consuming the latest causally available state before its query time. Materializing every action-prefix state is therefore unnecessary for the interface that \texttt{SeqFold} targets.

\begin{figure}[htbp]
    \centering
    \makebox[\linewidth]{\includegraphics[width=1.2\linewidth]{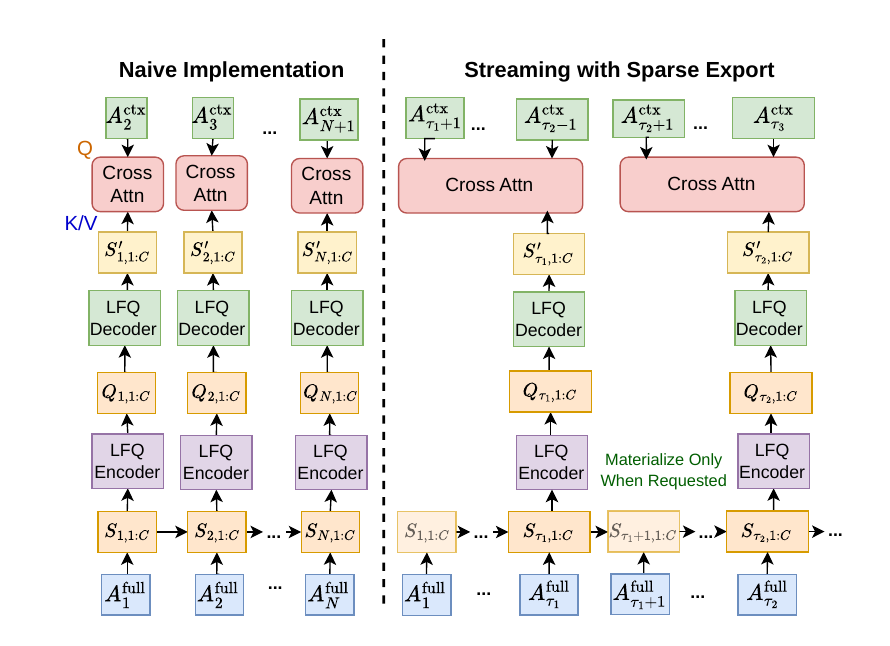}}
    \Description{A fused streaming export workflow. SeqFold scans actions in chronological order, updates the fusion recurrence after every action, and materializes compressed states only at requested timestamps. A grouped training example shows checkpoints reused by nearby queries.}
    \caption{Fused streaming compression with sparse timestamp export. SeqFold
    updates the fusion recurrence after every action, but materializes compressed
    states only at requested timestamps. The right panel illustrates the grouped
    training special case, where evenly spaced checkpoints $\tau_1,\tau_2,\ldots$
    are exported and queries between two consecutive checkpoints reuse the most
    recent exported state. More generally, the requested export set can be
    arbitrary, including backfill timestamps or the latest online
    serving timestamp.}
    \label{fig:scalable_seqfold}
\end{figure}

Compressor training likewise need not materialize the full prefix timeline. Training supervision is attached to downstream prediction records rather than to individual history actions, so each query needs only the latest exported state available before its effective query time. In one practical grouped-export variant with interval $G$, evenly spaced checkpoints are materialized, and queries between two consecutive checkpoints share the most recent exported state. Figure~\ref{fig:scalable_seqfold} illustrates this training-time pattern using checkpoints $\tau_1,\tau_2,\tau_3,\ldots$. This introduces only a bounded visibility delay while avoiding dense prefix-state materialization and enabling more efficient batching. The key systems distinction is therefore between state update and state materialization: the recurrence must still be evaluated after every action, but states need to be materialized only at requested timestamps.

\subsection{Fused Streaming Realization with Sparse Timestamp Export}
\label{sec:fused_streaming_export}
SeqFold addresses these bottlenecks by decoupling state update from state materialization.

\subsubsection{Streaming Kernel}
The summation in Equation~\ref{eq:seqfold_s} can be implemented as an equivalent streaming recurrence:
\begin{equation}
S_{i,c} = \alpha_{i,c} S_{i-1,c} + \beta_{i,c} A_i^{\mathrm{full}},
\label{eq:seqfold_recurrence}
\end{equation}
where $\alpha_{i,c}$ is the decay factor between consecutive actions and $\beta_{i,c}=\sigma(f_c(A_i^{\mathrm{full}}))$ is the action-dependent routing weight. The recurrence is evaluated after every action, so no historical action is skipped.

We implement this recurrence as a fused streaming kernel rather than as a sequence of standard framework tensor operators. Although Equation~\ref{eq:seqfold_recurrence} admits exact blockwise scan formulations, we prioritize a streaming design because our target workload requires sparse timestamp export without dense prefix-state materialization. The kernel performs the required chronological scan once along the action axis while exposing parallelism over channel--embedding tiles. Routing, temporal decay, state accumulation, and selected timestamp export are fused into one pass, avoiding both the dense $[B,N,C,D]$ prefix-state tensor and repeated memory traffic from materializing a separate state after every action. SeqFold therefore updates the running state after every action but materializes states only at requested timestamps. This separation is the systems counterpart of our timestamp-addressable abstraction: SeqFold preserves action-level causal evolution while exposing only the timestamped states that downstream records actually address.

For training, the backward pass follows the same streaming principle. Instead of saving the full $[B,N,C,D]$ prefix-state tensor, it stores final states and periodic FP32 checkpoints, then reconstructs local states through a reverse scan. The auxiliary workspaces still include channel-dependent gradient buffers, but the execution avoids materializing a full $C\times D$ fusion state for every action prefix.

\begin{algorithm}[t]
\small
\caption{Fused streaming SeqFold export}
\label{alg:fused_streaming_seqfold}
\begin{algorithmic}[1]
\Require Action states $\{A_i^{\mathrm{full}}\}_{i=1}^{N}$, action timestamps $\{t_i\}_{i=1}^{N}$, requested timestamp set $\mathcal{T}^{\mathrm{req}}_u$, decay factors $\alpha_{i,c}$, routing weights $\beta_{i,c}$
\Ensure Exported timestamp codes $\mathcal{Q}_u$
\State Initialize $S_c \gets 0$ for all $c=1:C$ and $\mathcal{Q}_u \gets \emptyset$
\For{$i=1$ to $N$}
    \For{$c=1$ to $C$}
        \State $S_c \gets \alpha_{i,c}S_c + \beta_{i,c}A_i^{\mathrm{full}}$
    \EndFor
    \For{each requested timestamp $\tau \in \mathcal{T}^{\mathrm{req}}_u$ such that $i=\max\{j:t_j<\tau\}$}
        \State $Q_{\tau,c} \gets \operatorname{LFQEncode}(\operatorname{RMSNorm}(S_c))$ for all $c=1:C$
        \State $\mathcal{Q}_u \gets \mathcal{Q}_u \cup \{Q_{\tau,c}\}_{c=1:C}$
    \EndFor
    \If{training and $i$ is a checkpoint position}
        \State Store an FP32 checkpoint of $S_c$ for the backward reverse scan
    \EndIf
\EndFor
\State \Return $\mathcal{Q}_u$
\end{algorithmic}
\end{algorithm}

\subsubsection{Timestamp Export Policies}
Formally, let $\mathcal{T}^{\mathrm{req}}_u$ be the set of target timestamps requested by downstream records for user $u$. For a requested timestamp $\tau \in \mathcal{T}^{\mathrm{req}}_u$, SeqFold exports the state after the latest action that strictly precedes $\tau$:
\begin{equation}
i(\tau) = \max\{i: t_i < \tau\}, \qquad
Q_{\tau,c} = \operatorname{LFQEncode}\left(
\operatorname{RMSNorm}\left(S_{i(\tau),c}\right)\right).
\label{eq:timestamp_export}
\end{equation}
This construction enforces the causal constraint required by backfill and online serving. A historical training record receives the user state available before its target event, while online serving can export the latest state at request time. Importantly, the number of requested states $V_u=|\mathcal{T}^{\mathrm{req}}_u|$ is determined by downstream data demand rather than by the length of the behavior sequence, allowing the same user history to support dense diagnostic export, sparse backfill, or single-state online serving without changing the recurrence. The continuous states materialized by the compressor scale as $O(V_u C D)$ before quantization and as $O(V_u C)$ after LFQ encoding, instead of $O(N C D)$ when every action prefix is exported.

Grouped training export is one special case of this interface and is the case illustrated in Figure~\ref{fig:scalable_seqfold}. When query timestamps are dense during compressor training, we can choose evenly spaced export checkpoints with interval $G$. In the figure notation, these checkpoints are denoted by $\tau_1,\tau_2,\ldots$, and queries between two consecutive checkpoints reuse the most recent exported state. In our implementation, we use zero-based action indexing, so the corresponding requested export set can be written as $\mathcal{T}^{\mathrm{req}}_u=\{G-1,2G-1,\ldots\}$, leading to:
\begin{equation}
\hat{A}_i^{\mathrm{ctx}} = \operatorname{Attn}\left(
    A_i^{\mathrm{ctx}},\operatorname{LFQDecode}\left(
    Q'_{\left\lfloor i/G\right\rfloor-1,1:C}\right)\right),
\label{eq:n_seqfold_attn}
\end{equation}
where actions are zero-indexed, $Q'_g$ denotes the code exported at the $g$-th group boundary, and the first group uses an empty initial state. The group size $G$ changes export granularity and training-state visibility, but it does not approximate the fusion recurrence: every action still updates the running state through Equation~\ref{eq:seqfold_recurrence}. For backfill and online serving, the same fused scan exports precisely the requested target timestamps in $\mathcal{T}^{\mathrm{req}}_u$.

The resulting implementation has three practical properties:
\begin{itemize}
\item \textbf{No dense prefix-state tensor}: The forward pass does not materialize the dense $[B,N,C,D]$ prefix-state tensor produced by a tensorized implementation. It keeps only tiled running fusion states during the scan and writes selected outputs.
\item \textbf{Output proportional to requested timestamps}: The number of materialized compressed states is controlled by the export set, ranging from every action prefix during diagnostic runs to sparse backfill timestamps or the latest online serving state.
\item \textbf{Controlled backward memory}: The backward pass stores final states and periodic FP32 checkpoints to reconstruct the reverse scan. Its auxiliary workspaces still contain channel-dependent buffers for accumulating gradients, but they avoid storing the full $C\times D$ fusion state for every action prefix.
\end{itemize}

\section{Experiments}

\subsection{Experimental Setup}

We evaluate SeqFold on proprietary datasets. Unless otherwise specified, the
main experiments use a four-layer backbone with hidden dimension 256, with the
compression module applied at every layer. In this main experimental setting,
each user sequence contains at most 2,000 actions, with an average length of
approximately 1,700. The scalability study benchmarks timestamp-state generation up to 10{,}000 actions. SeqFold compresses each layer into $C=64$ fusion
codes, each containing 64 LFQ bits, for a fixed storage cost of
\[
4\ \text{layers} \times 64\ \text{codes} \times 64\ \text{bits}
= 2{,}048\ \text{bytes}
\]
per user-state snapshot.

Throughout this section, storage denotes the serialized
feature payload required to describe one user at one target timestamp during
online serving or backfill. This timestamped definition is important
for downstream training: each historical record must be paired with the causal
user state available before its target event. In the main comparison with
recent raw-action baselines, each record consumes the latest available snapshot
before the target event rather than an online request-time state. This protocol
reflects downstream settings in which signals attached to the most recent
actions may be delayed or not yet finalized when features are materialized.
Storage does not include temporary checkpoints materialized during compressor
training. Total backfill volume still scales with the number of requested
timestamps for every method, while Table~\ref{tab:compression_baselines}
compares the payload size of one record.

Due to data confidentiality, we report
relative changes rather than absolute metric values. Unless otherwise
specified, the metric in this section is Query Precision-Recall AUC (PRAUC) on
the upstream action-prediction task used to train the compressor. Here, each
target event defines one query, and PR-AUC is computed at the query level and
then averaged across queries. The reference configuration is stated in each
table; positive values indicate
an improvement over that reference, whereas negative values indicate a
degradation. The downstream transfer subsection reports each task's own metric.

\subsection{Comparison with Compression Baselines}

\begin{table*}[t]
\centering
\caption{Comparison with full-sequence and sequence-compression baselines.
Performance is reported relative to SeqFold: Fusion + LFQ. Storage is measured
per user-state snapshot at one serving or backfill timestamp; values
marked ``avg.'' depend on the historical sequence length.}
\label{tab:compression_baselines}
\begin{tabular}{lrr}
\hline
\textbf{Model} & \textbf{Relative Metric Change} &
\textbf{Storage per Snapshot (Bytes)} \\
\hline
Full Action Sequence Model       & $+2.11\%$ & $6{,}963{,}200$ avg. \\
Fusion Model (without LFQ)       & $+0.72\%$ & $262{,}144$ \\
\textbf{SeqFold: Fusion + LFQ}   & $\mathbf{0.00\%}$ & $\mathbf{2{,}048}$ \\
32-Action Window Pooling         & $-1.10\%$ & $217{,}600$ avg. \\
64-Action Window Pooling         & $-1.38\%$ & $108{,}800$ avg. \\
Pooling to One Vector (Last Layer) & $-2.00\%$ & $1{,}024$ \\
Sample 64 Actions                & $-2.37\%$ & $262{,}144$ \\
Last 64 Actions                  & $-3.98\%$ & $262{,}144$ \\
\hline
\end{tabular}
\end{table*}

Table~\ref{tab:compression_baselines} compares SeqFold with the uncompressed
action sequence model and several compression strategies. The full model
retains every action representation from every backbone layer and therefore
provides an approximate upper bound on predictive quality. It improves the
evaluation metric by $2.11\%$ over SeqFold, but requires an average of
$6{,}963{,}200$ bytes per user-state snapshot. In contrast, SeqFold uses only $2{,}048$
bytes, corresponding to a $3{,}400\times$ reduction in storage. This result
shows that most of the predictive information in the multi-layer action
history can be retained without transmitting the complete continuous hidden
states.

The continuous Fusion Model isolates the effect of quantization. It uses the
same learned fusion mechanism as SeqFold but stores 64 dense 256-dimensional
vectors at every layer. Removing LFQ improves the metric by $0.72\%$, while
increasing storage from $2{,}048$ to $262{,}144$ bytes. Thus, LFQ provides an
additional $128\times$ reduction over continuous fusion for a $0.72\%$
relative metric decrease. This trade-off is important when the representation
must be stored and delivered to multiple downstream applications.

SeqFold also consistently outperforms compression methods that do not learn
which historical information should be preserved. Window pooling reduces the
sequence length by aggregating neighboring actions, but remains more than
$53\times$ larger than SeqFold and trails it by $1.10\%$--$1.38\%$. The
single-vector pooling baseline averages the action states from the final
backbone layer only and stores one 256-dimensional FP32 vector
($256 \times 4 = 1{,}024$ bytes). It therefore has a slightly smaller footprint
than SeqFold, but loses $2.00\%$, suggesting that one last-layer global vector
does not provide enough capacity to represent the diverse interests and
temporal patterns in a long user history.

Finally, retaining only 64 raw actions does not offer a competitive
quality--storage trade-off in the evaluated setting. Sampling 64 actions and
selecting the most recent 64 actions both require $262{,}144$ bytes---the same
storage as continuous fusion---yet underperform SeqFold by $2.37\%$ and
$3.98\%$, respectively. Sampling can discard sparse but informative behaviors,
whereas recent-window truncation relies more heavily on immediate recency.
Under the per-record snapshot protocol used here, this dependence becomes a
liability relative to a summary that aggregates information across the full
history. This result supports SeqFold in the timestamped export setting studied
here; performance may differ under other online serving assumptions. SeqFold
avoids both limitations by allowing all actions to contribute through learned,
channel-specific gating and time-aware decay.
Overall, the results indicate that the gain comes not only from reducing
sequence length, but from learning a structured multi-channel summary before
applying aggressive discrete quantization.

\subsection{Scalability of Timestamp-State Generation}

We isolate the timestamp-state generation operator from the backbone and
downstream decoders to measure the system-level effect of fused streaming. The
benchmark uses $C=64$ fusion channels, action dimension $D=256$, bfloat16
inputs, and forward--backward execution. Let $B$ denote batch size, $N$ denote
the per-user sequence length, and $V$ denote the total number of exported states
in a batch. We compare SeqFold's Triton fused streaming kernel with an optimized
tensorized PyTorch baseline that evaluates the same linear recurrence using
chunked cumulative products and cumulative sums. This baseline avoids
Python-level loops and represents a strong tensorized implementation of exact
blockwise scan using standard framework operators, but it still materializes a
state for every action prefix before selecting requested timestamps; in our
setting, each action update already spans a wide
$C\times D$ state ($64\times256$ in the main configuration), so we prioritize a
streaming design that directly supports sparse timestamp export rather than
additional sequence-axis scan orchestration. In this subsection, we benchmark
up to 10{,}000 actions with $B=10$ so that both implementations remain
runnable while keeping the same $C=64$, $D=256$, bfloat16, and
forward--backward protocol. Latency denotes operator execution time and memory
denotes peak allocated memory.

\begin{table}[t]
\centering
\caption{Scaling to long histories under dense export of all
action prefixes. We use $B=10$, $C=64$, $D=256$, bfloat16, and one
forward--backward pass. Memory is reported as fused/cumsum peak memory in GB.}
\label{tab:kernel_scalability_10k}
\begin{tabular}{rrrrr}
\hline
\textbf{$N$} & \textbf{Fused ms} & \textbf{Cumsum ms} &
\textbf{Speedup} & \textbf{Mem. GB} \\
\hline
2048  & 10.50 & 113.66  & $10.8\times$ & 7.2 / 9.6 \\
4096  & 20.97 & 262.77  & $12.5\times$ & 14.4 / 19.3 \\
8192  & 42.34 & 721.42  & $17.0\times$ & 28.7 / 38.6 \\
10000 & 51.77 & 1014.48 & $19.6\times$ & 35.0 / 47.1 \\
\hline
\end{tabular}
\end{table}

Table~\ref{tab:kernel_scalability_10k} extends the sequence-length study from
2K actions to long histories. After reducing batch size to keep
both implementations runnable, the fused kernel remains close to linear in
sequence length, whereas the cumsum baseline grows substantially faster. At
10{,}000 actions, fused streaming is $19.6\times$ faster and still uses less
peak memory, showing that timestamp-state generation remains practical even at
10{,}000 actions.

\begin{table}[t]
\centering
\caption{Effect of export density at 10{,}000 actions with batch size 10.
Memory is reported as fused/cumsum peak memory in GB.}
\label{tab:export_density_10k}
\begin{tabular}{rrrrr}
\hline
\textbf{$V/(BN)$} & \textbf{$V$} & \textbf{Fused ms} &
\textbf{Cumsum ms} & \textbf{Mem. GB} \\
\hline
1.0000 & 100{,}000 & 51.70 & 1017.15 & 35.0 / 47.1 \\
0.2500 & 25{,}000  & 26.44 & 971.16  & 9.9 / 21.9 \\
0.0625 & 6{,}250   & 19.98 & 966.45  & 3.6 / 19.9 \\
0.0156 & 1{,}560   & 18.45 & 963.29  & 2.0 / 19.8 \\
\hline
\end{tabular}
\end{table}

Table~\ref{tab:export_density_10k} fixes the action workload at 10{,}000
actions and varies only the number of requested timestamps. The fused kernel
must still scan the entire sequence, but sparse export sharply reduces its
materialization and backward costs: decreasing $V/(BN)$ from 1.0 to 0.0156
reduces latency from 51.70 ms to 18.45 ms and peak memory from 35.0 GB to 2.0
GB. In contrast, the cumsum baseline remains near 1 second and close to 20 GB
because it still constructs the full prefix-state timeline before selecting the
requested timestamps. At the sparsest setting, fused streaming is
$52.2\times$ faster and uses $9.9\times$ less peak memory, highlighting the
systems benefit of sparse timestamp-addressable export for long histories.

\subsection{Ablation Study}

\subsubsection{Number of Fusion Vectors}

\begin{table}[t]
\centering
\caption{Effect of the number of fusion vectors. The continuous Fusion Model
without LFQ is used to isolate the effect of fusion capacity, and the
64-vector configuration is used as the reference.}
\label{tab:compare_fusion_vector_num}
\begin{tabular}{lr}
\hline
\textbf{Number of Fusion Vectors} & \textbf{Relative Metric Change} \\
\hline
8 & -1.16\%  \\
16 & -1.13\% \\
32 & -0.72\% \\
\textbf{64} & \textbf{0.00\%} \\
128 & 0.08\% \\
\hline
\end{tabular}
\end{table}

The number of fusion vectors, $C$, controls the capacity of the compressed
sequence representation. As shown in
Table~\ref{tab:compare_fusion_vector_num}, using only 8 or 16 vectors degrades
the metric by $1.16\%$ and $1.13\%$, respectively. Increasing $C$ to 32
recovers a substantial fraction of this loss, reducing the gap to $0.72\%$.
The improvement continues up to 64 vectors, whereas doubling the capacity
from 64 to 128 yields only a marginal additional gain of $0.08\%$. This trend
indicates that very small fusion spaces create an information bottleneck, but
the benefit largely saturates beyond 64 vectors. We therefore choose
$C=64$ as the default, since it provides a favorable balance between
representation capacity, downstream attention cost, and storage.

\subsubsection{Number of Fusion Layers}

\begin{table}[t]
\centering
\caption{Effect of applying fusion to different numbers of backbone layers.
The one-layer configuration uses only the last layer, while the four-layer
configuration uses all backbone layers and serves as the reference.}
\label{tab:compare_fusion_layer_num}
\begin{tabular}{lr}
\hline
\textbf{Number of Fusion Layers} & \textbf{Relative Metric Change} \\
\hline
1 & -0.97\% \\
2 & -0.52\% \\
3 & -0.44\% \\
\textbf{4} & \textbf{0.00\%} \\
\hline
\end{tabular}
\end{table}

Table~\ref{tab:compare_fusion_layer_num} studies whether it is sufficient to
compress only the final backbone layer. Using the last layer alone trails the
full four-layer configuration by $0.97\%$. Adding the second and third fusion
layers progressively narrows the gap to $0.52\%$ and $0.44\%$, while applying
fusion at all four layers achieves the best result. In the proposed
architecture, each fusion module summarizes the hidden states available at a
different depth of the backbone and is consumed by the corresponding
downstream cross-attention layer. The result suggests that compressed
representations from different depths provide complementary signals to the
downstream model. Based on this observation, we apply SeqFold at every
backbone layer in the default configuration.




\subsubsection{Temporal Decay}

\begin{table}[t]
\centering
\caption{Effect of temporal decay. Performance is reported relative to the
variant without temporal decay. A constant factor specifies the remaining
weight after one year.}
\label{tab:temporal_decay}
\begin{tabular}{lr}
\hline
\textbf{Temporal Decay} &
\textbf{Relative Metric Change} \\
\hline
No Time Decay & 0.00\% \\
Constant (0.3 after one year) & -0.02\% \\
Constant (0.6 after one year) & 0.03\% \\
Learned Decay & 0.18\% \\
\hline
\end{tabular}
\end{table}

Table~\ref{tab:temporal_decay} compares fixed and learned temporal weighting.
A fixed one-year retention factor of 0.3 slightly reduces the metric by
$0.02\%$, while a factor of 0.6 yields only a $0.03\%$ improvement. In
contrast, learning the decay rates improves the metric by $0.18\%$ over the
no-decay variant. These results indicate that a single hand-tuned decay
schedule is too restrictive: different fusion channels benefit from adapting
their temporal horizons to the data. We therefore use channel-specific learned
decay rates in SeqFold.

\subsubsection{LFQ Quantization Capacity}

We next vary the capacity of the LFQ representation while keeping the fusion
architecture fixed. This experiment measures how aggressively the continuous
fusion state can be discretized before quantization becomes the dominant
information bottleneck.

\begin{table}[t]
\centering
\caption{Effect of the LFQ code length. Performance is relative to the default
64-bit configuration. Storage is measured per user-state snapshot.}
\label{tab:compare_lfq_capacity}
\begin{tabular}{lrr}
\hline
\textbf{Bits per Code} & \textbf{Storage (Bytes)} &
\textbf{Relative Change} \\
\hline
16 & 512 & -0.99\%\\
32 & 1{,}024 & -0.36\% \\
\textbf{64} & \textbf{2{,}048} & \textbf{0.00\%} \\
96 & 3{,}072 & -0.05\% \\
\hline
\end{tabular}
\end{table}

Table~\ref{tab:compare_lfq_capacity} shows a clear quality--storage trade-off.
Reducing the code length from 64 to 16 bits lowers the per-snapshot footprint
from 2,048 to 512 bytes, but incurs a $0.99\%$ performance loss, confirming that
overly aggressive quantization removes useful information from the learned
fusion vectors. A 32-bit code recovers most of this loss while retaining a
$2\times$ storage reduction relative to the default, trailing the 64-bit code
by only $0.36\%$. Increasing the code length beyond 64 bits provides no further
benefit: the 96-bit configuration consumes $50\%$ more storage and performs
slightly worse by $0.05\%$. The difference between 64 and 96 bits is small,
and there is no measured benefit from the larger code. We therefore use
64-bit LFQ codes as the default operating point.

\subsection{Downstream Transfer Across Decoder Dimensions}

We next evaluate whether the compact LFQ codes transfer to downstream search
and query autocomplete (QAC) models under different decoder capacities.
Table~\ref{tab:downstream_transfer} reports relative changes in the primary
metrics for both tasks as the projection dimension varies from 32 to 128.
Relative changes are measured against task-specific baselines that do not use
an exported SeqFold user-state embedding as input.

\begin{table}[t]
\centering
\caption{Transfer of SeqFold codes on search and QAC across projection dimensions.}
\label{tab:downstream_transfer}
\begin{tabular}{llr}
\hline
\textbf{Task} & \textbf{Projection Dimension} & \textbf{Relative Change} \\
\hline
Search & 32 & 0.2433\% \\
Search & 64 & 0.4908\% \\
Search & 128 & 0.6149\% \\
QAC & 32 & 0.0966\% \\
QAC & 64 & 0.5311\% \\
QAC & 128 & 0.5954\% \\
\hline
\end{tabular}
\end{table}

Table~\ref{tab:downstream_transfer} shows positive gains for both search and
QAC at every tested projection dimension. On search, the gain increases from
$0.24\%$ at dimension 32 to $0.61\%$ at dimension 128. QAC shows a similar
pattern, improving from $0.10\%$ at dimension 32 to $0.60\%$ at dimension 128,
with most of the gain already realized by dimension 64. These results suggest
that the LFQ codes remain useful across downstream capacity budgets, allowing
each application to choose its decoder dimension without changing the shared
upstream exported representation.

\section{Conclusion}
\label{sec:conclusion}

We presented \texttt{SeqFold}, a framework for compressing multi-layer,
long-horizon action representations into compact discrete codes. Its learned
channel routing and time-aware fusion combine semantic and temporal signals,
while LFQ provides a compact interface that downstream applications can decode
into task-specific representation spaces. A fused streaming implementation
supports long-sequence training by maintaining a bounded continuous state and
batching attention over periodically exported checkpoints.

Experiments on proprietary datasets show that SeqFold reduces the average
representation size by $3{,}400\times$ relative to the full action sequence
while remaining within a 2.11\% relative metric gap. It provides a better
quality--storage trade-off than the evaluated pooling, sampling, and
truncation baselines; the fused streaming kernel scales to long histories and
benefits from sparse timestamp export; and the learned codes transfer to
downstream search and query autocomplete across decoder capacities. Ablations
further show that multi-channel and multi-layer fusion matter, learned
temporal decay outperforms fixed decay, and gains saturate around 64
channels and 64 bits per code. These results support SeqFold as a shared
interface between large user sequence models and downstream applications that
consume compact timestamp-aligned states; future work will study additional
downstream tasks, adaptive checkpoint intervals, and task-aware compression
budgets.

\section{Ethical Considerations}
The experiments in this paper use proprietary datasets of user actions and
report only aggregate results. We do not include user-identifying examples or
release raw behavioral data in the manuscript. Because compact exported user
states can still encode sensitive behavioral signals, practical deployments of
SeqFold should follow existing privacy, retention, and access-control policies
for user representations. In addition, downstream ranking or recommendation
models built on user histories may inherit or amplify historical biases present
in the underlying data. We therefore view SeqFold as a systems interface for
more efficient state reuse, not as a substitute for application-level auditing,
fairness evaluation, or governance before production deployment.

\bibliographystyle{ACM-Reference-Format}
\bibliography{references}

\end{document}